\documentclass[runningheads]{llncs}
\usepackage[T1]{fontenc}
\usepackage{graphicx}
\usepackage{booktabs}
\usepackage[misc]{ifsym}

\usepackage{hyperref}
\usepackage{booktabs}   
\usepackage{caption}
\usepackage{subcaption}
\usepackage{makecell}
\usepackage{multirow}

\usepackage{amsmath,amsfonts,bm}

\def\eqref#1{equation~\ref{#1}}

\def\1{\bm{1}}

\DeclareMathAlphabet{\mathsfit}{\encodingdefault}{\sfdefault}{m}{sl}
\SetMathAlphabet{\mathsfit}{bold}{\encodingdefault}{\sfdefault}{bx}{n}

\def\gO{{\mathcal{O}}}

\def\sX{{\mathbb{X}}}

\newcommand{\E}{\mathbb{E}}

\newcommand{\softmax}{\mathrm{softmax}}

\usepackage[capitalize,noabbrev]{cleveref}

\usepackage{pifont} 

\usepackage{listings}
\usepackage{xcolor} 

\definecolor{comment}{HTML}{787b80}
\definecolor{keywords}{HTML}{fa8d3e}
\definecolor{string}{HTML}{86b300}

\usepackage{algorithm}
\usepackage{algorithmic}

\newcommand{\sigmatitle}{\texorpdfstring{$\sigma$}{sigma}}

\begin{document}
\title{\sigmatitle-GPTs: A New Approach to Autoregressive Models}
\titlerunning{\sigmatitle-GPTs: A New Approach to Autoregressive Models}

\author{ Arnaud Pannatier\inst{1,2} \and Evann Courdier\inst{1,2} \and Fran\c cois Fleuret\inst{3}}
\authorrunning{A. Pannatier et al.}

\institute{Idiap Research Institute, Martigny, Switzerland \and
Ecole Polytechnique Fédérale de Lausanne, Lausanne, Switzerland \and
Université de Genève, Geneva, Switzerland}

\maketitle
\begin{abstract}
  Autoregressive models, such as the GPT family, use a fixed order, usually left-to-right, to generate sequences. However, this is not a necessity. In this paper, we challenge this assumption and show that by simply adding a positional encoding for the output, this order can be modulated on-the-fly per-sample which offers key advantageous properties. It allows for the sampling of and conditioning on arbitrary subsets of tokens, and it also allows sampling in one shot multiple tokens dynamically according to a rejection strategy, leading to a sub-linear number of model evaluations. We evaluate our method across various domains, including language modeling, path-solving, and aircraft vertical rate prediction, decreasing the number of steps required for generation by an order of magnitude
  \setcounter{footnote}{0} 
  \footnote{The code of this work is available at \url{https://github.com/idiap/sigma-gpt}}.

\end{abstract}




\section{Introduction}

Transformers demonstrate exceptional autoregressive capabilities across modalities. The traditional take for autoregression is to follow the natural order of the data, for example, left-to-right for text. In the case of vision, the usual scheme is to unfold the images following a raster-scan order and to use transformers to model the obtained sequence. In this work, we make a distinction between the order of the input data and the order of autoregression, highlighting that while they are typically aligned in most applications, they need not be. Our investigation involves training and generating sequences in a randomly shuffled order using transformers. While changing the sequence order is more challenging during training, it also reveals fascinating properties of the models.

\begin{figure}
  \centering
  \includegraphics[width=1.0\textwidth]{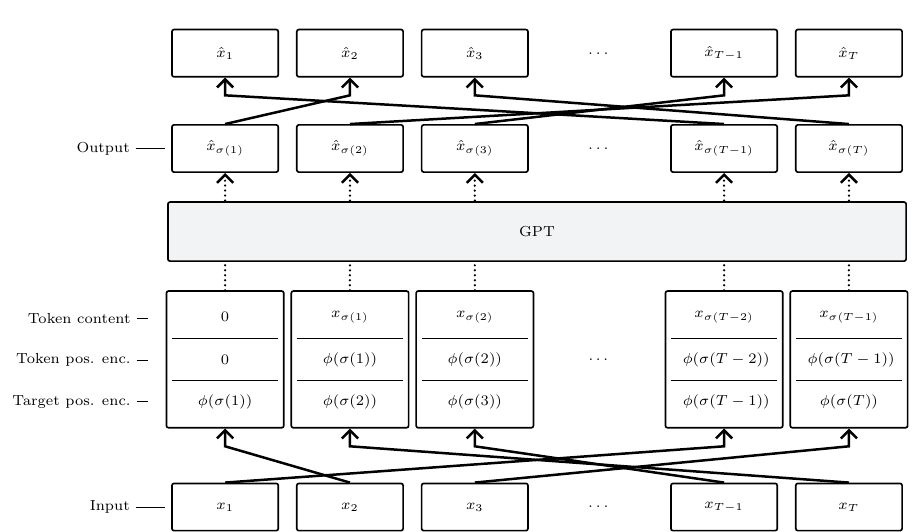}
  \caption{
    In our $\sigma$-GPT, an arbitrary shuffling order $\sigma$ can be chosen on-the-fly for every sample. It induces an input order $0,\sigma(1), \sigma(2), \dots$ and an output order $\sigma(1),\sigma(2), \sigma(3), \dots$, where the input is first padded with a $0$ to ensure a consistent number of tokens. Tokens are shuffled accordingly, and these orders are both encoded separately with two positional encodings concatenated to the input, allowing the model to sample consistently in the autoregressive process. The output is finally shuffled back to the true order.}
\end{figure}

\newcommand{\ok}{\ding{52}}
\newcommand{\x}{\ding{55}}
\begin{table}
  \centering
  \setlength{\tabcolsep}{12pt}
  \begin{tabular}{lccc}
                                 & \bf{$\sigma$-GPT} & \bf{GPT} & \makecell{\bf{Diffusion} \\ \bf{Models}}  \\
    \toprule
    \bf{Sample Anywhere}         & \ok               & \x       & \x                       \\
    \bf{Conditional Density Estimation} & \ok & \x & \x \\
    \bf{Arbitrary Conditioning}  & \ok               & \ok      & \Large\hspace{-2mm}\textbf{ \texttildelow}     \\
    \bf{Infilling}               & \ok               & \x       & \Large\hspace{-2mm}\textbf{ \texttildelow}                      \\
    \bf{Burst-Sampling}          & \ok               & \x       & \ok                      \\
    \bf{Log-likelihood Training} & \ok               & \ok      & \x                       \\
    \bottomrule
  \end{tabular}
  \caption{Comparison between our approach, a standard causal transformer encoder (called GPT here), and diffusion models. Our model allows the sampling of a token at any position in the sequence, to model the remaining density according to a partially sampled sequence, naturally supports infilling, and can be used to sample the sequence by burst allowing faster generation. Compared to diffusion models, it can be trained easily using cross-entropy.}
  \label{tab-comparison}
\end{table}

By breaking away from the standard autoregression order, one can use the model to predict the tokens in any particular order. With this scheme, the model is capable of predicting at any moment of the generation the conditional distribution of the remaining tokens. Having these estimates allows quantifying the possible outcomes of the generation at any given point. More interestingly, they can be leveraged to do rejection sampling, allowing to generate sequences by burst with a dynamical number of steps.

This work is structured as follows, we first introduce $\sigma$-GPTs and shuffled autoregression, and show that a model trained with this method combined with a curriculum method can even increase the performance of the underlying model. We then present the additional properties of $\sigma$-GPTs, summarized in \Cref{tab-comparison}, in particular for estimating conditional probabilities and we present our token-based rejection sampling scheme which allows for generating the sequence per burst and its theoretical properties. We evaluate our model and our scheme on three main tasks, which are open text generation, path-solving, and aircraft vertical rate prediction.

\paragraph{Contributions:}
\begin{itemize}
  \item Introduce $\sigma$-GPT, a novel architecture, with two positional encodings related respectively to the input and output order, that allows a causal transformer to generate sequences in any order which can be modulated on the fly for any pass through the model.
  \item Demonstrate that our method can reach similar performance as left-to-right trained autoregressive models when trained with a curriculum scheme.
  \item Demonstrate that our method can be used to generate samples in any order, allowing for the generation of samples conditioned on any part of the sequence.
  \item Introduce a novel token-based rejection sampling scheme that leads to the generation of samples per burst.
\end{itemize}

\section{Methodology}
\subsection{\sigmatitle-GPTs: Shuffled Autoregression}

We propose a novel approach for training autoregressive models, which involves doing next-token prediction on a shuffled input sequence.
We present $\sigma$-GPT, where $\sigma$ denotes the permutation used to shuffle the sequence, and by GPT we mean any causal transformer encoder (or causal transformer decoder without cross-attention) such as~\cite{radford2019gpt2}. To train such a model, each sequence is shuffled randomly during training. The model is then tasked to predict the next token in the shuffled sequence conditioned on all the tokens it has seen before. This training is done as usual with a standard cross-entropy loss. Besides the randomization of the order of the sequence and the addition of a double positional encoding, no other changes are needed to the model or training pipelines.
For the rest of the paper, we use `left-to-right order` to mention the usual order in which models are trained, even in the case of 2D data which are usually mapped to a sequence using a raster-scan order. And we use `random order` to mean that the input has been shuffled.

\subsection{Double Positional Encodings}

To be able to model sequences in any order, each token needs to have information about its position and the one of the next token in the shuffled sequence.
Specifically, when handling a sequence of tokens alongside a given permutation $\sigma$, every token contains three distinct pieces of information: its value $x_{\sigma(t)}$, its current position $\sigma(t)$, and the position $\sigma(t+1)$ of the subsequent token in the shuffled sequence, that are all concatenated. The necessity for double positional encoding arises from the intrinsic characteristics of transformers. Given that each token attends to every previous token in a position-invariant manner, each token needs to contain information about its position in the original sequence, so other tokens can know where they are located. And each token needs to know the position of the next token in the shuffled sequence as it is the target of the prediction. The double positional encoding is the only architectural change needed to train autoregressive models in random order. In this work, we used the standard sinusoidal positional encoding~\cite{vaswani2017attention} for both the input and output positional encodings.

\subsection{Conditional Probabilities and Infilling}

\begin{figure}[htbp]
  \centering
  \includegraphics[width=\columnwidth]{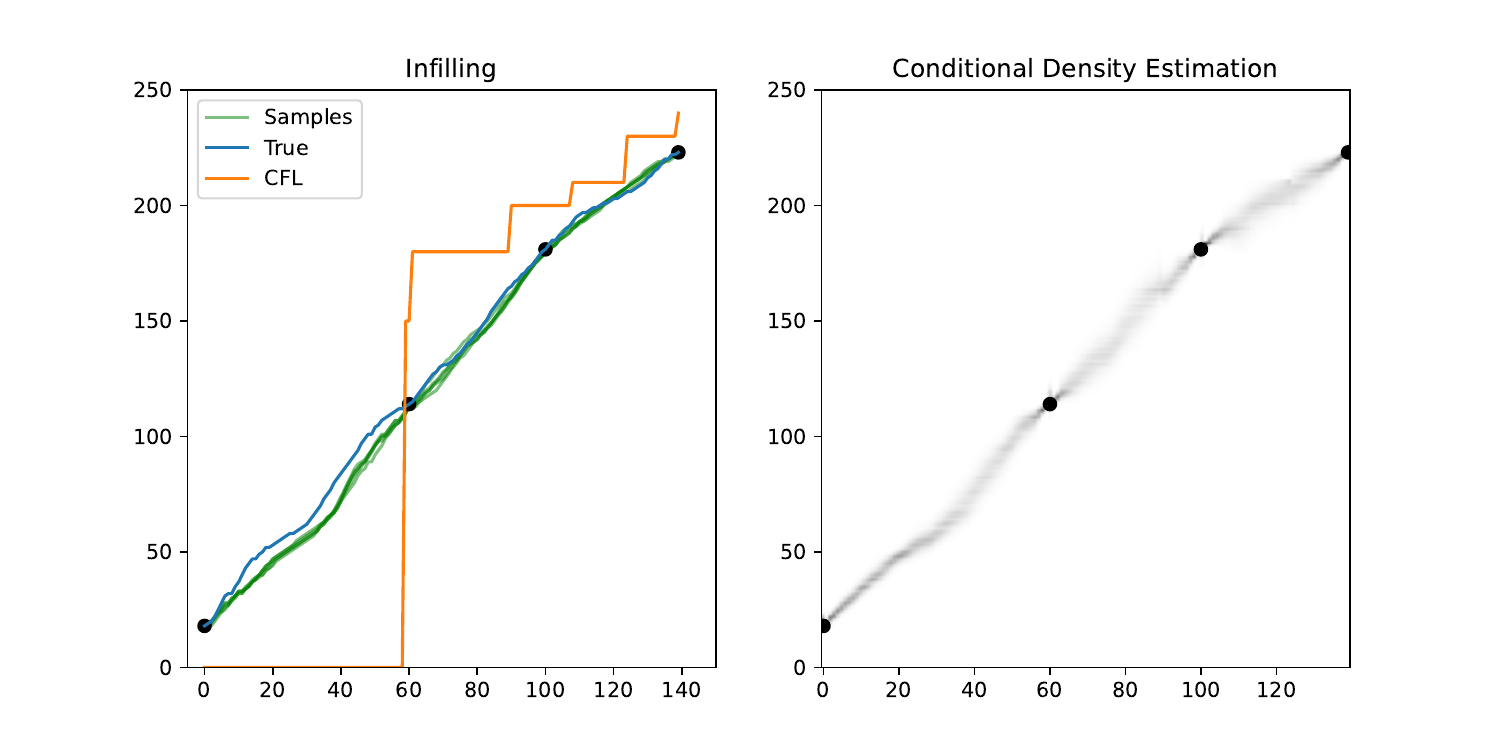}
  \caption{(Left.) We can infill the sequence by conditioning on the known part (black points). (Right.) We can also have estimates of the density at any point of the sequence.}
  \label{fig:climb-infill-condpost}
\end{figure}

\begin{figure*}
  \centering
  \begin{subfigure}[t]{0.49\textwidth}
    \includegraphics[width=\textwidth]{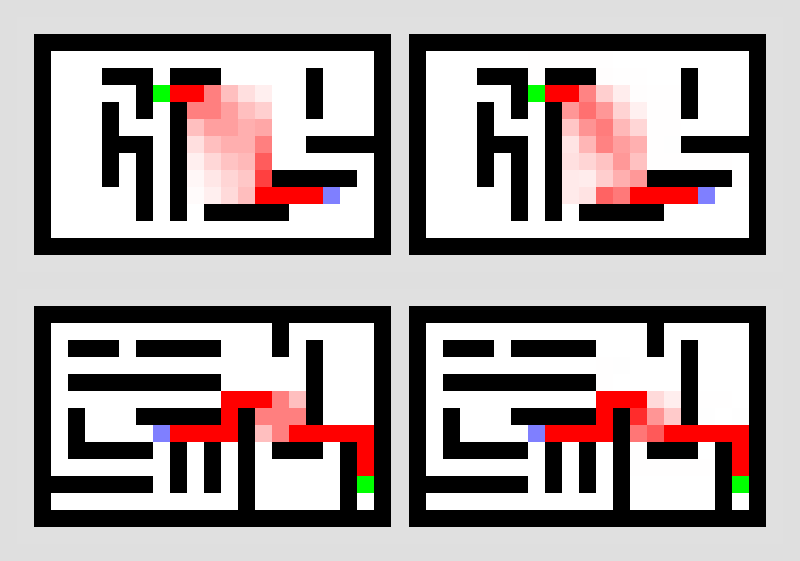}
    \caption{(Left.) The theoretical density of the optimal path in the maze. (Right.) The estimated probability of the class `path` at every position before starting autoregression. We see that the model has good estimates of the true density.}
    \label{fig:density-maze}
\end{subfigure}
\hfill
\begin{subfigure}[t]{0.49\textwidth}
    \includegraphics[width=\textwidth]{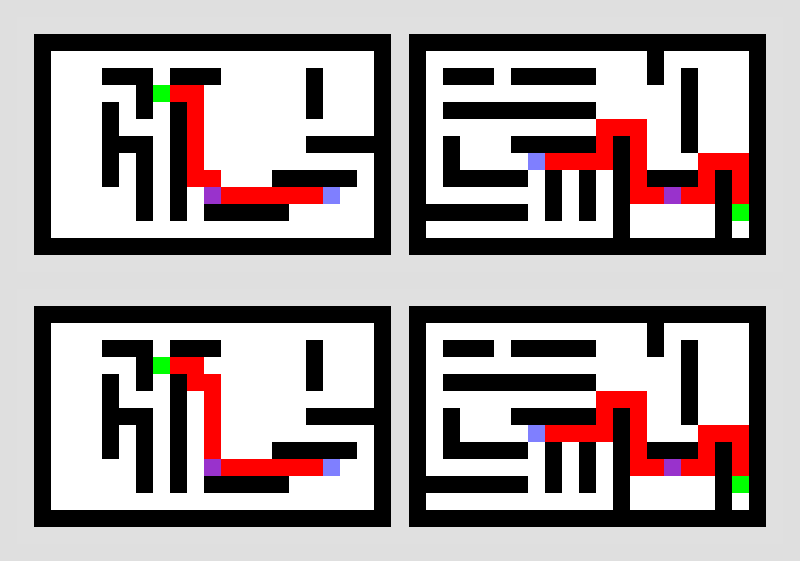}
    \caption{Two different conditional samplings for each maze. The known part of the path (purple) is prompted first, and the rest of the sequence can be completed coherently.}
  \end{subfigure}
  \label{fig:sampling-maze}
  \caption{Conditional density estimation and infilling on the maze path-solving task.}
\end{figure*}

Our method allows making conditional density estimation of the rest of the sequence. It is capable of making predictions all over the task space conditioned on any known subpart of the task. This can be done by prompting the model with the known part of the sequence and then decoding, in parallel and in one pass, the remaining tokens. Such evaluations are not possible with autoregressive models trained in a left-to-right order, as they need to follow the specific order they've been trained in. Examples, showing that the model usually has good estimates of the unconditioned distribution, can be seen in \Cref{fig:climb-infill-condpost,fig:density-maze}.

Directly related to conditional density estimation, is that our method naturally supports infilling, as it is straightforward to prompt the model with the known part of a signal and to decode auto-regressively or by burst the rest of the signal. \Cref{fig:climb-infill-condpost,fig:density-maze} shows example of such samplings.

\subsection{Token-based Rejection Sampling}

\newcommand{\xst}{x_{\sigma_{<t}}}
\newcommand{\p}{p(x_t | \sX)}
\newcommand{\q}{q(x_t | \sX, \xst)}
\begin{algorithm}[ht]
  \caption{Token-based rejection sampling, following notation of \cite{chen2023speculativesampling}}
  \label{algo:burst-rejection-sampling}
  \begin{algorithmic}
    \STATE Given minimum target length $T$, y trained $\sigma$-GPT, and number of orders $N_o$
    \STATE Given a prompt {$x_i \in \sX$} of length $t_0$ of initial tokens. ($\sX$ can be the empty set)
    \STATE Set $t = t_0$
    \WHILE {$t<T$}
    \STATE In parallel, compute distribution conditioned on prompt $p(x_i| \sX),\forall i \in t,\dots,T$
    \STATE In parallel, sample at every position $\tilde{x}_i \sim p(x_i| \sX), \forall i \in t,\dots,T$
    \STATE Draw $N_o$ random order $\sigma$ and in parallel, compute all logits $q(x_i | \sX, \tilde{x}_{\sigma_{<i}}), \forall i \in t, \dots, T$
    \STATE In parallel sample $T - t$ variables $u_i \sim U[0,1], \forall i \in t,\dots,T$ from a uniform distribution.
    \STATE In parallel, compute the acceptance decision $a_i = u_i < \min \left(1, \frac{q\left(\tilde{x}_i | \sX, \tilde{x}_{\sigma_{<i}}\right)}{p(\tilde{x}_i | \sX)}\right)$ for every order.
    \STATE Select the order that accepts the most tokens before seeing a first rejection.
    \STATE Keep that order and add the $a$ accepted tokens before the first rejection to the prompt.
    \STATE Set $t = t + a$
    \ENDWHILE
  \end{algorithmic}
\end{algorithm}

Autoregressive generation is a slow process as each token has to be generated sequentially. Even with caching strategies, this still scales linearly with the sequence length and it becomes prohibitively expensive for long sequences~\cite{villegas2023phenaki}. As our model allows for the generation of tokens in any order, we can leverage that fact and sample tokens in parallel at every position of the sequence. We can then evaluate the candidate sequence under different orders and accept multiple tokens in one pass. This algorithm runs efficiently on GPU as both the sampling at every position and the evaluation under different orders can be made in parallel, in a forward pass, and using an adapted KV-caching mechanism. We describe this caching mechanism more in detail in~\Cref{kvcaching} of the supplementary material. When conditioned on partially completed sequences the model outputs distributions that are compatible with different possible outcomes, and when evaluating under different orders for generation, the distribution of tokens is constrained to tokens that are compatible with the previous tokens seen in one given order. As both the sampling and evaluation can be done in parallel, we can compute the acceptance decision efficiently for every token.

This strategy outputs a decision for each remaining token, but the decisions made by models become sometimes nonsensical when two mutually exclusive tokens are part of the prompt. Once a rejection is seen, all subsequent accepted tokens in the order of evaluation should be discarded. Indeed, the scheme rejects tokens that are incoherent with the ones already seen, and asking a model to make predictions based on incoherent tokens might lead to incoherent decisions. Using multiple orders allows keeping the one that accepts the most tokens in its evaluation. Even if it is dynamic, this algorithm can still easily generate multiple samples at once, by accepting the same amount of tokens for each sequence in the batch. Our rejection sampling algorithm is given in pseudo-code in \Cref{algo:burst-rejection-sampling}.

Other models such as Mask Git~\cite{chang2022maskgit} or diffusion models~\cite{ho2020ddpm,austin2021d3pm} are doing generation by burst. However, these models usually require fixing the number of steps or a masking schedule beforehand. Our method on the other hand adapts dynamically to the underlying statistics of the data and thus does not require this extra hyper-parameter. We evaluate it on three synthetic cases to showcase this dynamic capability, we present the results in~\Cref{ssec:burst-sampling}.




\subsection{Other Orders} \label{sec:otherorders}
Our double positional encoding scheme allows for training and evaluating models in any order.
Using a randomized order during training allows conditional density estimation, infilling, and burst-sampling at inference time.
However the double positional encoding scheme allows any order to be used, and it can be used to train models in a deterministic order that is not left-to-right.
As an example, we use a deterministic `fractal` order to see how it compares to a random or left-to-right order.
This order starts in the middle of the sequence then recursively goes to the first quarter and three-quarters of the sequence, and goes on recursively until all the positions have been visited. Such an order is fully deterministic, yet we make the hypothesis that this order leads to more difficult training for the model as it cannot rely on the locality of the information. We present the results in~\Cref{sec:fractal}. Note that under perfect models, the order of modeling and decoding should not matter because of the chain rules of probability. We give more details about it in \Cref{chain-rule} of the supplementary material.

\subsection{Denoising Diffusion Models}
Denoising diffusion models~\cite{ho2020ddpm} is a family of generative models that can also be used to generate sequences in a few steps. They are trained to reverse a diffusion process that is applied to the data. Diffusion processes can be both continuous and discrete. In this work, we use as a baseline only the discrete diffusion case, in particular using a uniform diffusion process~\cite{austin2021d3pm}. To be able to compare the methods fairly, we use the same transformer architecture for both $\sigma$-GPT and the diffusion model, changing only the training objective. Compared to $\sigma$-GPT, diffusion models are not dynamic and require a fixed number of steps to generate a sequence, independently of the underlying statistics of the data. They also don't natively support conditional density estimation and infilling. 

\section{Results}
\subsection{General performance}
We tested our model across three main distinct tasks: language modeling, maze path solving, and aircraft vertical-rate prediction.

\begin{itemize}
  \item Language Modeling: We used both the GPT-2 (123M) model on the Wikitext-103 dataset~\cite{merity2017wikitext103} and GPT-2 (345M) on OpenWeb Text~\cite{gokaslan2019openwebtext}.
  \item Maze Path Solving: This task involves determining a valid path between a starting and ending point in 13 x 21 mazes featuring 15 barriers. Presented with an image of an empty maze with start and end points, the model is tasked with producing an image with a legitimate path.
  \item Aircraft Vertical-Rate Prediction: This task uses real aircraft trajectory data, with its aircraft type. The data represents trajectories conditioned by air traffic control directives. The model's objective is to predict the vertical trajectory from a plane's current altitude to a specified control level.
\end{itemize}
Additionally to these tasks, we created a synthetic benchmark for evaluating our burst-sampling algorithm.
\begin{itemize}
  \item Product Dataset: This toy example represents a pure product law case and is made of a sequence of length 100 with two classes (0,1) given by a Bernoulli law with p = 10\%.
  \item Step Dataset: This toy example comprises sequences of two classes (0,1) of length 100 which are 0 everywhere except on a step of length 10 placed randomly in the sequence
  \item Joint Law dataset: This toy example represents a pure joint law and consists of a sequence of length 100 with 100 different classes, the model should predict a random generation of these different classes.
\end{itemize}

The general results of our models are presented in~\Cref{tab:results}. These results indicate that training in a random order while requiring more compute-time as we describe in~\Cref{sec:training-efficiency}, reaches similar performances to left-to-right trained models. 
For the text modeling, to have a fair comparison during training, we monitor the validation perplexity of the sequence evaluated in a left-to-right order. Training in random order for text modeling was plateauing at a higher left-to-right validation perplexity, but using a curriculum scheme allows reaching the same performances, as presented in \Cref{sec:curriculum-learning}.
For the path solving and the vertical rate prediction, the models were able to reach the same left-to-right validation loss during training. In inference, we noticed a one percent drop in accuracy compared to diffusion models and left-to-right trained GPT. For the vertical rate prediction task, the dataset that we used is limited to around 23.000 different sequences, we noticed that the standard left-to-right GPT was sometimes stuck repeating the same altitude, we think this is a modeling issue due to the small data regime. $\sigma$-GPT does not seem to suffer as much from this problem and offers a decrease in MSE.  
We hypothesize that this behavior comes from using a random order in inference which forces the model to fix some tokens over the whole sequence early in the generation. By doing so, the model gains the advantage of having a sketch of the whole sample and then concentrates on completing a coherent sample. 

\begin{table}
  \caption{
    General results. We report the validation perplexity for text generation, the test accuracy for the maze solver, and the mean squared error (MSE) for the vertical rate prediction. $\sigma$-GPT reaches a similar performance as GPT in text generation and maze solving and it outperforms GPT in the case of the vertical rate prediction. We report the validation perplexity for the text generation. For the path solver, we report the test accuracy on 1000 novel mazes. For the vertical prediction task, we report the mean squared error on the test set. We report the mean and standard deviation for the path-solving and the vertical rate prediction task. We do not report the validation error for the text generation for the discrete diffusion (Dis. Diff) as the training objective is different.}
  \label{tab:results}
  \centering
  \begin{tabular}{lccrr}
    \toprule
                      & \multicolumn{2}{c}{\bf{Text-generation}} & \bf{Path Solving} & \bf{Vertical Rate} \\                      
                     & OWT Val Perp. ($\downarrow$) &  Wiki-103  Val Perp. ($\downarrow$)  & Accuracy ($\uparrow$) & MSE ($\downarrow$) \\
    \midrule
    \bf{GPT}          & 18.14                          & 20.30 & 99.60 $\pm$ 0.70 & 274.8 $\pm$ 70.7 \\
    \bf{$\sigma$-GPT} & 18.64                          & 16.69 & 98.30 $\pm$ 0.67 & 141.4 $\pm$ 4.1  \\
    \bf{Dis. Diff.}   & -                              & -     & 99.20 $\pm$ 0.67 & 105.94 $\pm$ 1.3 \\
    \bottomrule
  \end{tabular}
\end{table}

\subsection{Training Efficiency} \label{sec:training-efficiency}

\begin{table*}
  \caption{
    Training efficiency. Number of steps/epochs required to reach the same performance and comparison with as GPT trained causally. As learning to predict in any order is a more challenging task, it is expected to need more computing time to reach the same accuracy.
    We don't report the standard deviation for text generation as we limited the training to one run.
  }
  \label{tab:trainingefficiency}
  \centering
  \begin{tabular}{lrrr}
    \toprule
    \bf{Order}        & \bf{Text-generation} & \bf{Maze Solver} & \bf{Climbing Rate} \\
    \midrule
    \bf{$\sigma$-GPT} & 32500                & 78.0 $\pm$ 6.5   & 110.7 $\pm$ 4.5    \\
    \bf{GPT}          & 16500                & 19.3 $\pm$ 4.9   & 25.0 $\pm$ 3.6     \\
    \bottomrule
  \end{tabular}
\end{table*}
Modeling sequences in a random order is a more challenging task than modeling in left-to-right order. 
We think this is due to two main factors, at the beginning of the sequences models cannot rely on adjacent tokens to make educated guesses for the next token. Second some tasks are harder to learn in one direction than another and by modeling the data in any direction, we are always in the harder scenario. We give an example of one task that is harder to learn in one direction in~\Cref{multinomial-rw} of the supplementary material. 

This implies that we expect and see an increase in the number of steps or epochs required to learn a task. As previously mentioned, we don't see experimentally a drop in the validation performance of our model in the case of the path-finding algorithm or the vertical rate forecasting, but the time to reach the same performance increased. In the case of text modeling, the models plateaued before reaching the same accuracy when trained in random order. We treat that case in the following section.
We report in table \Cref{tab:trainingefficiency}, the increase in training steps or epoch to reach the same accuracy. We see that most of the time, the number of epochs or steps needed to reach the same performance drastically increases. We think again that this is due to the increased complexity of modeling the sequence without having to rely on local information.

\subsection{Curriculum Learning} \label{sec:curriculum-learning}
\begin{table}[ht]
  \caption{
    Curriculum learning. We monitor the Validation Perplexity using a left-to-right order during training to have a comparable evaluation. We see that there is a gap between the model trained purely in a left-to-right fashion (GPT) and others trained in a random order ($\sigma$-GPT). Training for longer and larger models didn't help in removing that gap. We introduce a curriculum learning scheme that starts presenting the model with some percentage of the data (written in the corresponding label) in a left-to-right order at the beginning of the training and goes linearly to 100\% of sequence in a random order at the end of the training. We see that training with this scheme removes the gap between $\sigma$-GPT and regular GPT and it reaches even better than left-to-right performance in the WikiText-103 case.
  }
  \label{tab:curriculum}
  \centering
  \begin{tabular}{lr}
    \toprule
                                   & \bf{Text-generation} Val Perp. ($\downarrow$) \\
                                   & Min. Left-to-Right                      \\
    \midrule
    \multicolumn{2}{c}{\it{Openweb Text} - GPT (345 M)}                            \\
    \midrule
    \bf{GPT}                       & 18.14                                         \\
    \bf{$\sigma$-GPT curr. 50\% }  & 18.64                                         \\
    \bf{$\sigma$-GPT no curr.}     & 30.43                                         \\\midrule
    \multicolumn{2}{c}{\it{WikiText 103 - GPT (128M)} }                            \\ \midrule
    \bf{$\sigma$-GPT curr. 50\%}   & 16.69                                         \\
    \bf{$\sigma$-GPT curr. 100\% } & 19.38                                         \\
    \bf{$\sigma$-GPT curr. 10\%}   & 19.45                                         \\
    \bf{GPT}                       & 20.30                                         \\
    \bf{$\sigma$-GPT no curr.}     & 39.85                                         \\
    \bottomrule
  \end{tabular}
\end{table}
For text modeling, we found a gap in validation perplexity in the left-to-right order between models trained purely in a random order and models trained in a left-to-right order. We see in~\Cref{tab:curriculum} that $\sigma$-GPT is stuck at larger perplexity in both Open Web Text and WikiText-103 (30.43 vs 18.14 and 39.85 vs 20.30). We found that training for longer and using larger model didn't help in reducing that gap.
To solve that problem, we introduced a curriculum learning scheme where the model is shown first more sequences in left-to-right order and progressively learns to model the sequence randomly. Surprisingly using this scheme helped drastically the model which managed to get even better performance than left-to-right trained transformers in the Wikitext-103 case and reduce drastically the gap for models trained on OpenWebText.


\subsection{Open Text Generation: t-SNE of Generated Sequences}

\begin{figure}
  \begin{subfigure}[t]{0.24\textwidth}
    \centering
    \includegraphics[width=\textwidth]{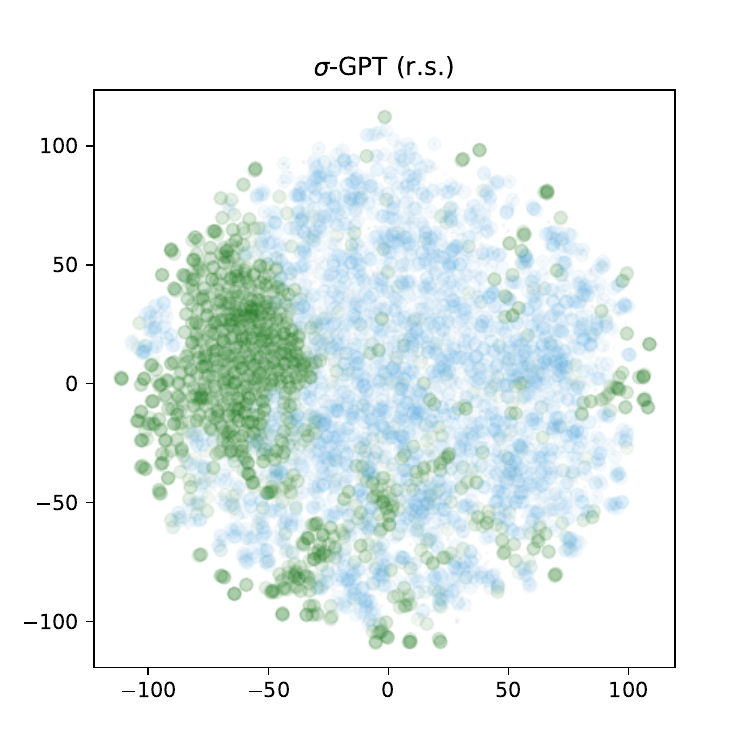}
    \caption{Rej. Sampling}
    \label{fig:tsne-rejection}
  \end{subfigure}
  \begin{subfigure}[t]{0.24\textwidth}
    \centering
    \includegraphics[width=\textwidth]{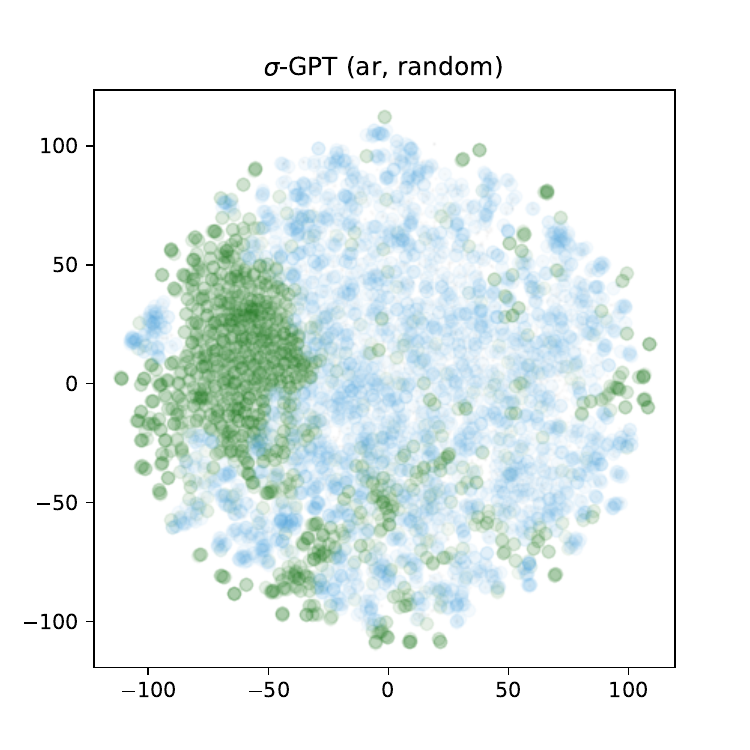}
    \caption{$\sigma$-GPT}
    \label{fig:tsne-sigma-gpt}
  \end{subfigure}
  \begin{subfigure}[t]{0.24\textwidth}
    \centering
    \includegraphics[width=\textwidth]{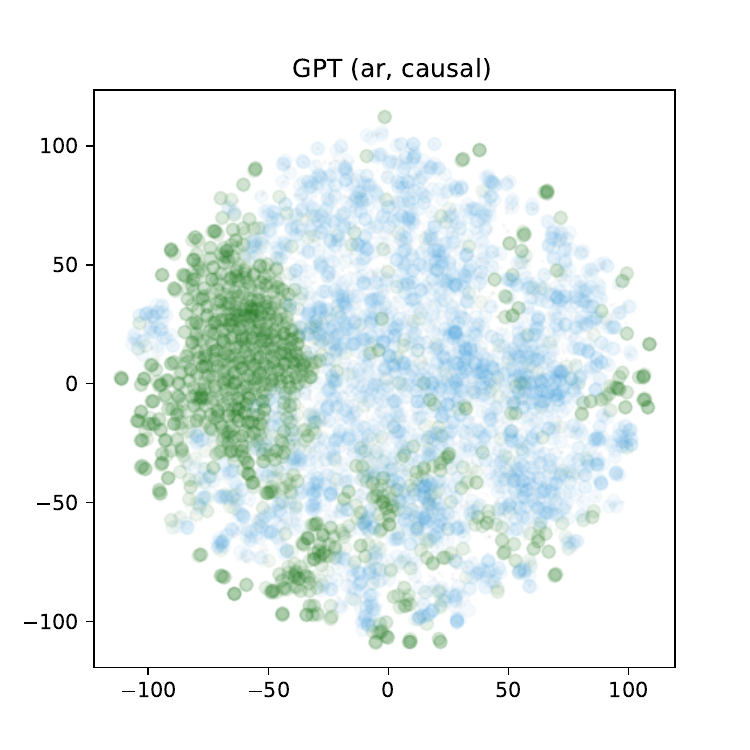}
    \caption{GPT}
    \label{fig:tsne-gpt}
  \end{subfigure}
  \begin{subfigure}[t]{0.24\textwidth}
    \centering
    \includegraphics[width=\textwidth]{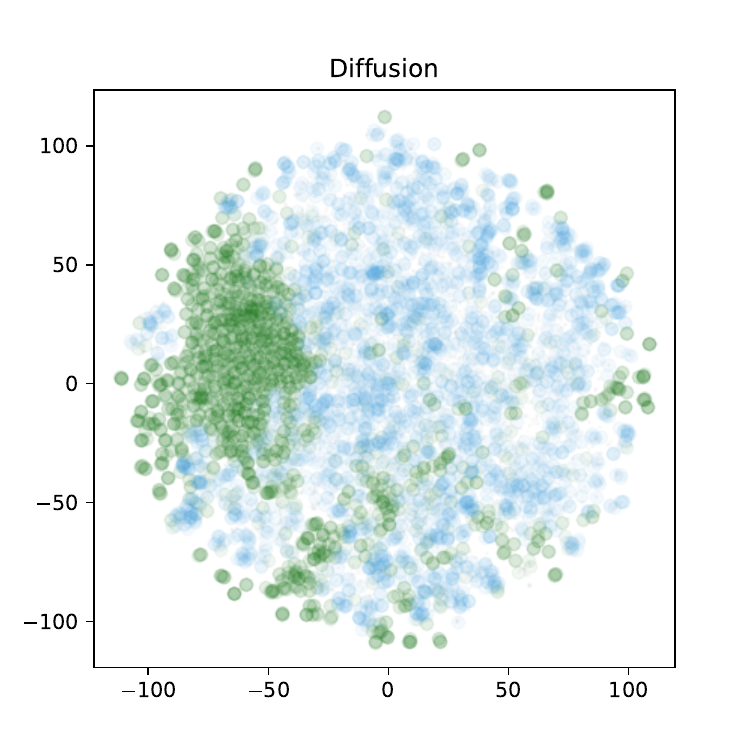}
    \caption{Diffusion}
    \label{fig:tsne-diffusion}
  \end{subfigure}
  \caption{2D t-SNE of \texttt{text-small-3-embeddings} of 3000 sequences generated by each method. We compute the t-SNE of all the embeddings together, and then we display in each graph the embeddings of the validation set (green), the embeddings of the corresponding method (blue), and the embeddings of the other methods (gray). 
  We see that the embeddings of the generated sequences have the same overall distribution compared to validation sets, which seems to indicate that $GPT$, $\sigma$-GPT, $\sigma$-GPT with burst-sampling, and diffusion models can generate sequences of similar quality.}
  \label{fig:t-sne}
\end{figure}

To get a qualitative sense of the generated text by the different methods, We generate 3000 sequences of 1024 tokens with each method, embed each sequence using an embedding model, and then project the embeddings to 2D using t-SNE. We present the results in~\Cref{fig:t-sne}. We used Open-AI \texttt{text-embedding-3-small}~\cite{openai2024textembedding3small} to embed the generated sequences into a single 1536 vector embedding.
We represent as green embeddings of sequences of the validation set, used as reference. We compute the t-SNE using the whole 15'000 embeddings and then plot each method (blue) and the other considered method (small gray dots). We first see that embeddings of GPT, $\sigma$-GPT, $\sigma$-GPT with burst-sampling, and diffusion are spread over the whole space, showing that the model can generate sequences that are coherent with the validation set. 

\subsection{Training and Generating in Fractal Order} \label{sec:fractal}

We describe here the results that we get when training a GPT using a deterministic, but not left-to-right order. We described the order in~\Cref{sec:otherorders}. We train a GPT using this specific order for the different tasks and present the results in~\Cref{tab:fractal}. We found that training in that order was as difficult for the model as training in a random order, and we noticed a small drop in performance compared to $\sigma$-GPT. We suspect that this is due to the high discontinuity of the order of the sequence, which is such that two consecutive tokens are seen far away in the sequences. When predicting the first tokens, the model therefore cannot rely on information contained in neighboring tokens to make its prediction. As the training behavior seen in models trained in random and fractal order is similar, we think that the drop in training efficiency comes more from the fact that the model cannot exploit this neighboring information than changing the order at every batch.

Additionally, models trained in a fractal cannot be used as such for infilling and conditional density estimation and therefore cannot be used with our rejection sampling scheme. As the order is fixed for every batch, it might not even need to have a double positional encoding.

\begin{table}
  \centering
  \begin{tabular}{lccc}
    \toprule
    & \bf{Text-generation} & \bf{Path Solver} & \bf{Vertical Rate} \\
    & Val Perp. ($\downarrow$), Rand. ord. & Test Acc ($\uparrow$) &  MSE ($\downarrow$)  \\
    \midrule
    \bf{$\sigma$-GPT} &  24.46                & 98.30 $\pm$ 0.67 & 141.4 $\pm$ 4.1    \\
    \bf{Fractal GPT}  &  27.79                    & 98.00 $\pm$ 1.94 & 145.7 $\pm$ 2.6    \\
    \bottomrule
  \end{tabular}
  \caption{Results for GPT trained in a fractal order compared to a standard left-to-right (GPT) and random ($\sigma$-GPT) order. We found that training models in this highly non-continuous order is as hard as training them in a random order, and additionally, models trained in that order cannot be used for conditional density estimation, infilling, or rejection sampling. For text modeling, we report the model perplexity on the validation set, in a random order for $\sigma$-GPT and in fractal order for the fractal GPT, as left-to-right validation perplexity is meaningless for fractal GPT which did not see sequences in that order during training.}
  \label{tab:fractal}
\end{table}

\subsection{Memorizing} \label{sec:memorizing}
\begin{figure}
  \centering
  \includegraphics[width=0.7\textwidth]{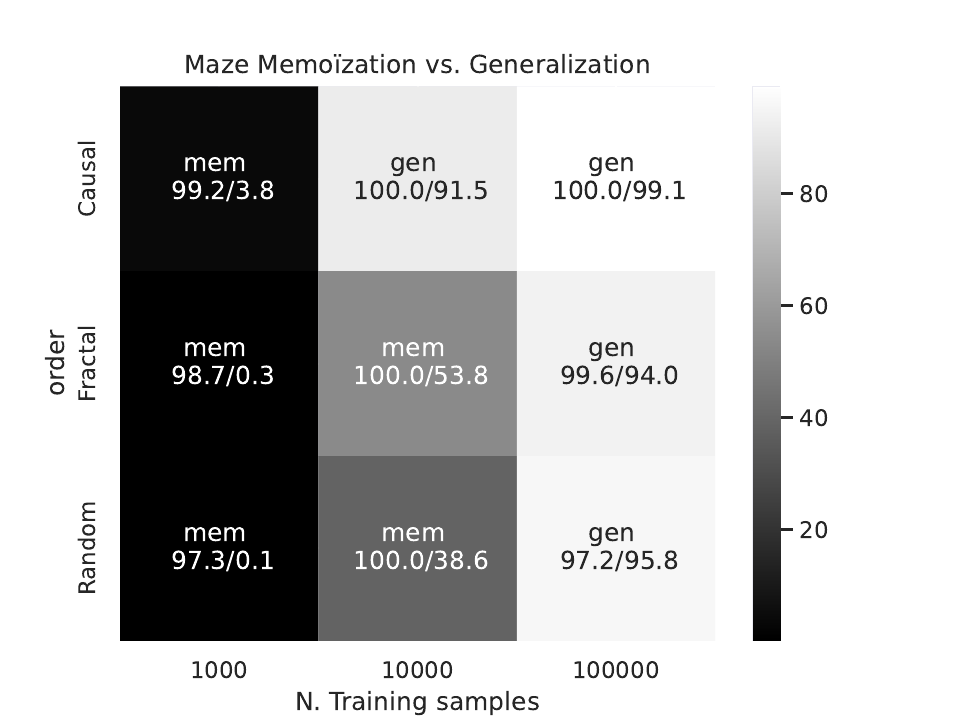}
  \caption{Number of examples needed to switch from memorization to generalization. The model is trained on a restricted dataset size in the path-finding task. We see that the model trained in a random order needs more examples to switch from memorization to generalization. At 1k samples both models are fully in a memorization regime, at 100k both generalize but in between, at 10k, the model trained in a random order is still in a memorization regime.}
  \label{fig:memorization}
\end{figure}

As learning sequences in any direction is harder than modeling them under a predefined order, we also expect that the critical dataset size when the model switches from memorization to generalization will increase. We follow the same hypotheses than~\cite{varma2023explaining}, namely that the model has two mechanisms, one generalizing and one memorizing the data. As the mechanism of generalizing is more efficient as the dataset grows it will be selected by gradient descent once the size of the dataset gets beyond a critical size. As learning in a random order is a more difficult task, we expect that generalization is more difficult in that setup as well, hence the memorization regime should hold for bigger dataset sizes. We reduce drastically the training dataset size in the case of the path-finding task and we present the results in~\Cref{fig:memorization}. Once it gets to 1000 examples both models trained in left-to-right, fractal, and random order are in a memorizing regime, getting perfect accuracy on the training data but very low on the validation data. Conversely, once the dataset gets bigger than 100k examples models trained in all the different orders are in a generalization regime. The transition happens in between and we find that it happens faster in the left-to-right order: at around 10k samples, the models trained left-to-right can generalize, while models trained in a random order are still in a memorization regime. We see also that models trained in a fractal order start generalizing faster than models trained in a random order, suggesting that the model can rely on seeing always the same order to generalize more rapidly.

\subsection{Infilling and Conditional Density Estimation}

We show in~\Cref{fig:climb-infill-condpost} that our model can be used to infill the sequence by conditioning on the known part of the sequence. In this figure, the larger points are part of the prompt and one can see the generated sequence complete sequence that matches the prompt.
This figure also shows that our model has good estimates of the density at any point of the sequence. We represent at each point in the sequence the probability of the next sample given the known part of the sequence as shades of gray, the darker the more probable. We see that during generation, the model sees multiple possible outcomes that are coherent with the known part of the sequence. They are then constrained to a single sequence during the sampling. For left-to-right trained models, we can only have estimates of the density for the next tokens and we can't know what the model estimates for the rest of the sequence.
In the case of the path-finding task, we show in \Cref{fig:density-maze} that the model conditioned only on an empty maze has good estimates of the true density of the optimal paths, highlighting that the model has already partially solved the problem before starting generation. During sampling, the order of the generation and the sampling procedure influence which path is selected from this joint law.
We show as well in~\Cref{fig:sampling-maze}, that we can constrain the generation of mazes and that the model can generate coherent samples based on some prompted tokens that can be chosen on the fly.

We also give some interactive examples of text generation in the supplementary material.

\subsection{Token-based Rejection Sampling Scheme} \label{ssec:burst-sampling}

\begin{figure*}
  \centering
  \begin{subfigure}{0.3\textwidth}
    \centering
    \includegraphics[width=\textwidth]{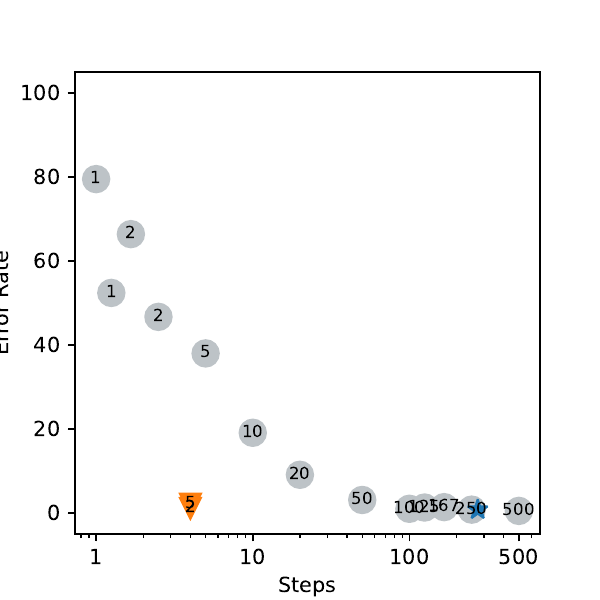}
    \caption{Path-Solving}
    \label{fig:burst-maze}
  \end{subfigure}
  \begin{subfigure}{0.3\textwidth}
    \centering
    \includegraphics[width=\textwidth]{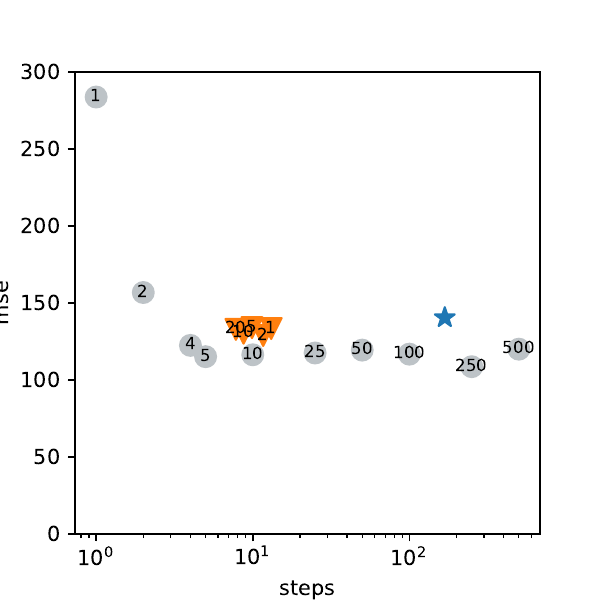}
    \caption{Vertical-Rate Pred.}
    \label{fig:burst-climbingrate}
  \end{subfigure}
  \begin{subfigure}{0.3\textwidth}
    \centering
    \includegraphics[width=\textwidth]{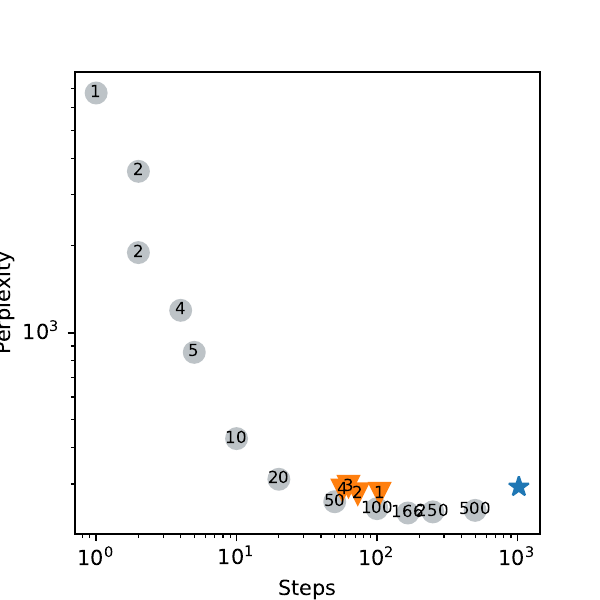}
    \caption{Text Modeling}
    \label{fig:burst-text}
  \end{subfigure}
  \begin{subfigure}{0.3\textwidth}
    \centering
    \includegraphics[width=\textwidth]{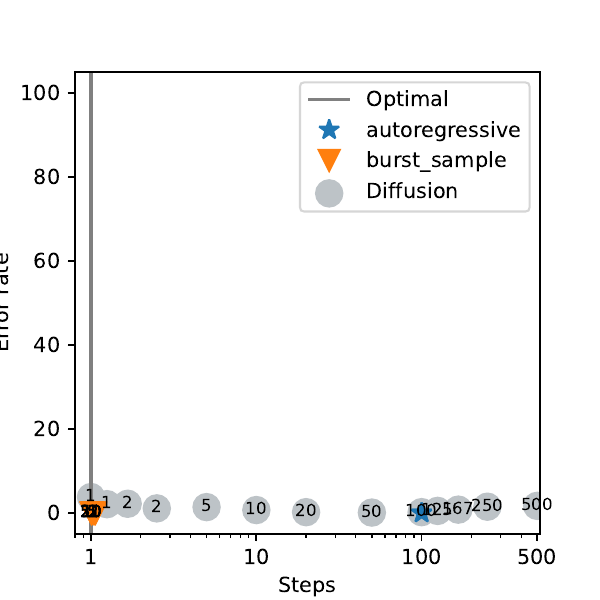}
    \caption{Product Task}
    \label{fig:burst-product}
  \end{subfigure}
  \begin{subfigure}{0.3\textwidth}
    \centering
    \includegraphics[width=\textwidth]{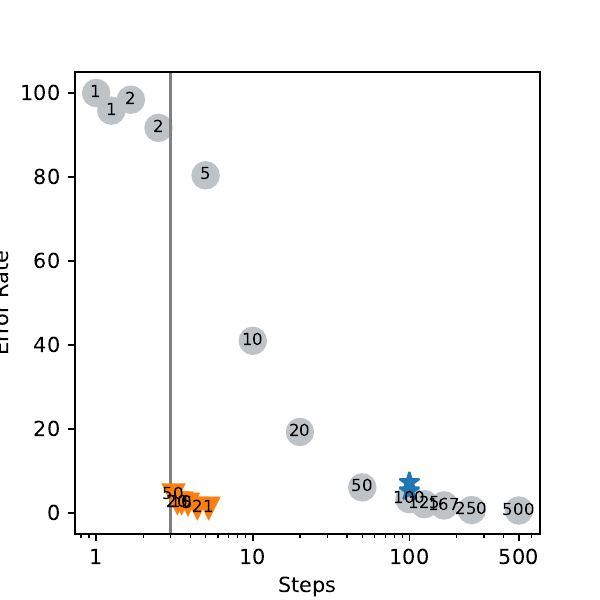}
    \caption{Step Task}
    \label{fig:burst-step}
  \end{subfigure}
  \begin{subfigure}{0.3\textwidth}
    \centering
    \includegraphics[width=\textwidth]{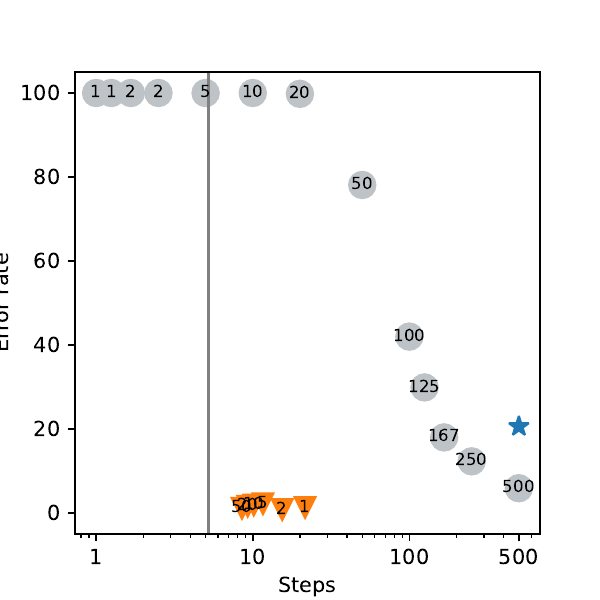}
    \caption{Permutation Task}
    \label{fig:burst-permutation}
  \end{subfigure}
  \caption{We plot the performance vs steps of our $\sigma$-GPT used for autoregression in random order (blue), to $\sigma$-GPT with rejection sampling per burst (orange) against diffusion models (gray). We denote in the text the predefined number of steps chosen for generation in the diffusion models. For rejection sampling, we note the number of orders used for the evaluation. We see that increasing the number of orders leads to a decreased number of steps. For the synthetic tasks, we also represent heuristics for the optimal number of steps needed to generate the sequence (gray line), as described in~\Cref{burst-sampling} of the supplementary material, and we see that our scheme is close to this heuristics.}
\end{figure*}

We applied our token-based rejection sampling scheme to the problem of text generation, path-finding, and vertical rate forecasting \Cref{fig:burst-maze,fig:burst-climbingrate,fig:burst-text}. We found that in the three cases, our method was able to generate samples of comparable quality with an order of magnitude fewer steps than the autoregressive method. We compared to discrete diffusion models as well, and we can see that our method always outputs coherent samples, while the samples generated by the diffusion model are sometimes incoherent if the number of steps is not high enough. In the case of path-solving and in the three synthetic tasks, we see that at a comparable number of steps for both methods, our model shows better generation quality.

We tested our token-based rejection sampling scheme on three synthetic cases. We estimate the number of steps required to generate the sequence using a perfect model in~\Cref{burst-sampling} of the supplementary material. We found that our scheme was close to the optimal heuristics in all three cases. In the case of the product dataset, requiring only one step to accept the sequence. In the case of the step dataset, our scheme required four steps to accept the sequence. In the case of the joint law dataset, our scheme required a few more steps, which is expected as the task is more complicated. We see that it manages to generate valid samples with a number of steps close to the optimal heuristics and with a large increase in performance compared to diffusion models at the same step. 


\section{Related works}

\paragraph{Shuffling in Language Models:}

The objective of $\sigma$-GPT is to model shuffled sequences. XLNet~\cite{yang2019xlnet} uses a similar objective and shuffling of the sequence order as a pretraining task for sentence encoding in the context of natural language understanding. 
While both approaches are modeling shuffled sequences, they still differ in their implementations: we use a double positional encoding and a regular causal mask instead of masking based on two streams and the modification of the attention matrix. The two models differ as well in their applications, XLNet is used to encode information similarly to BERT while our approach is generative. A recent approach by \cite{golovneva2024reverse}, trains a transformer both on a left-to-right and right-to-left order to solve a classic problem of the transformer model to understand tokens relations in both directions. In a similar setting, it has been shown that text corpora have a preferred left-to-right order of generation~\cite{papadopoulos2024arrows} and that training in reverse order can lead to a decrease in performance. This is a possible explanation of why training $\sigma$-GPT on text generation needed extra care compared to other tasks. In vision, the raster-scan order is mainly used to unfold 2D patches into sequences. However, a recent work by~\cite{kakogeorgiou2024spot} showed that using a set of predefined orders of unfolding the images can improve performance. They pass the order information not by using a double positional encoding, but by a fixed number of beginning-of-sequences tokens.

\paragraph{Burst-Sampling Scheme:}
Other works are trying to solve the problem of the linear time required by autoregression using burst-sampling. Maskgit~\cite{chang2022maskgit}, for example, uses a BERT-like Masked Language Model (MLM) and a custom decoding scheme, which samples multiple tokens at the same time to generate an output. The number of tokens generated at each pass is fixed by a masking schedule and a confidence-driven decoding scheme is used to choose which tokens to predict next.
Another approach, \cite{lezama2022tokencritic} relies on an auxiliary model to guide the generation process. Alternatively, the approach of \cite{lee2022draftandrevise} focuses on generating preliminary drafts of an image, which are then iteratively improved. Current Video generation methods~\cite{chang2023muse,villegas2023phenaki} are leveraging a MaskGit-like approach~\cite{chang2022maskgit} for generation because autoregressive generation of video frames would be too costly. Our rejection sampling scheme allows to generate the sequence by burst but in contrast to other schemes, the number of tokens accepted is dynamic and depends on the data being modeled. This allows for faster generation when the underlying data distribution is simple.

\paragraph{Discrete Diffusion Models:}

Diffusion models are also able to generate sequences in a few steps. We compared our approach with a discrete diffusion baseline~\cite{austin2021d3pm}. In the original work, discrete diffusion was also used to generate text, however without the possibility of conditional density estimation and infilling. Most of the diffusion approaches do not support conditional density estimation and infilling by default. Consistency Models~\cite{song2023consistency}, which be adapted from continuous diffusion models, can be used to infill images but in a continuous case. However, there have been recent discrete text-diffusion models that allow infilling~\cite{lou2024discrete,gulrajani2023conddiff}.

\section{Conclusion}

Training GPT-like models in different orders offer different desirable properties. It allows for the conditional prediction based on any subset of the tokens of the sequence, it naturally can be used for infilling, and as the model can do partial prediction, we can leverage them to do rejection sampling and accept multiple tokens at the same time during generation. Our findings indicate that conditional prediction learned by the models matches the theoretical partial distribution showing that the model is indeed able to understand and reconstruct the signal in any order. As the training objective of modeling sequences in any order is harder than training in a fixed order, it has an impact on the training efficiency, and in a small dataset size, we show that it leads to more memoïzation. Finally, we showed that our model was able to generate sequences by burst using a novel per-token rejection sampling scheme, reaching optimal heuristics in some cases and decreasing the number of steps needed for generation by an order of magnitude.

\paragraph{Acknowledgement:}
We thank Youssef Saied for his help and good remarks on the overall project. We thank Romain Fournier for his precious help on some theoretical aspects of the analysis.
Arnaud Pannatier was supported by SkySoft ATM for the project ``MALAT: Machine
Learning for Air Traffic'' and the Swiss Innovation Agency
Innosuisse under grant number 32432.1 IP-ICT. -- Evann Courdier was supported by the “Swiss Center for Drones and Robotics
- SCDR” of the Swiss Department of Defence, Civil Protection and Sport via
armasuisse S+T under project No 050-38.

\bibliographystyle{splncs04}
\bibliography{main}

\appendix

\section{Shuffled sequences are harder to learn} \label{multinomial-rw}

Using a random order in the training of a transformer often leads to a harder task to learn.
To showcase this property, we designed the following task. The model is asked to model a lazy random walk that starts from a multinomial distribution made of four Dirac distributions (at 100, 120, 130, 140), with equal probabilities. The random walk evolves with the following law:

\begin{align}
  p(X_t | X_{t - 1}) = \begin{cases} \frac{2}{5} & \text{if} \; \vert X_t - X_{t-1} \vert = 1 \\
              \frac{1}{5} & \text{if} \; X_t = X_{t-1}                 \\
              0           & \text{otherwise}
                       \end{cases}
\end{align}

With $p(X_0=100)=p(X_0=120)=p(X_0=130)=p(X_0=140)=1/4$.

In this setup predicting future tokens conditioned on past tokens is easy and is given by:
\begin{equation}
  X_{t_3} \sim \mathcal{N}(X_{t_2}, (t_3 - t)\sqrt{0.8})
\end{equation}
with $t_1 < t_2 < t_3$

note: $\mathbb{E}(X^2) = 2* 0.4 * 1 + 0.2*0  = 0.8$

The exact probability mass function (pmf) can also be computed:
\begin{equation}
  p(X_{t_3} = x | X_{t_2} ) = \sum_{o = 0}^n \binom{n}{o} \binom{n - o}{u} p_u^u p_d^d p_o^o
\end{equation}
with $n = t_3 -t_2$, $u = n-o+x-X_{t_2}$, $d = n - o - u$.

Predicting tokens conditioned on future information is much harder in that setup as the model has to take into account the initial multinomial distribution and keep track of all the different possibilities.

In such cases the exact pmf is given by :

\begin{equation}
  p(X_{t_1} = x | X_{t_2}) = \frac{ (\sum_{i \in \mathcal{S}} P(i,x,t_1) ) P(x, X_{t_2}, t_2 - t_1) }{ \sum_{i \in S} P(i, X_{t_2}, t_2) }
\end{equation}
Where $P(a,b,n) = \# \text{paths } a \rightarrow b \text{ in } n \text{ steps } = \sum_{o = 0}^n \binom{n}{o}\binom{n - o}{u}$. with $u = n-o+b-a$.

\begin{figure}[htbp]
  \centering
  \includegraphics[width=\columnwidth]{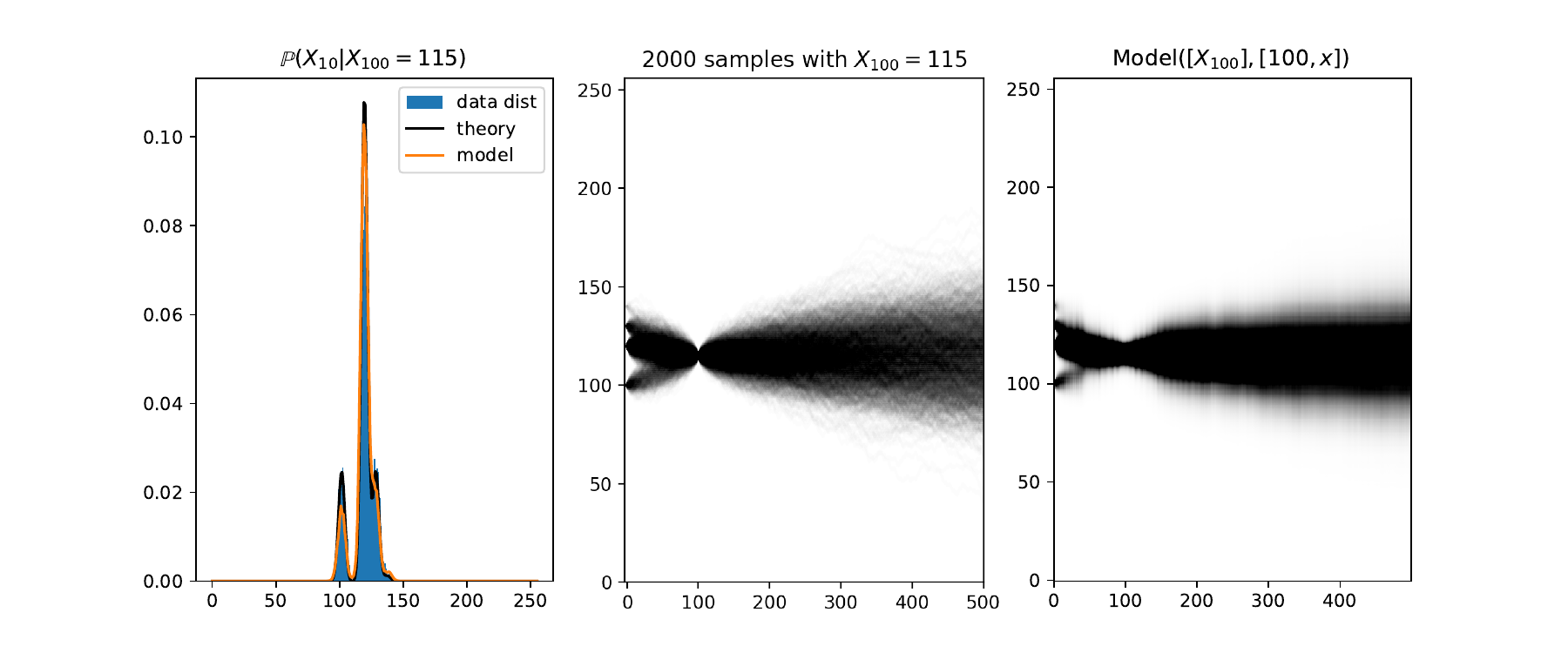}
  \caption{Random walks at different altitudes results. (\textbf{Left}) Comparison with the theoretical density at position 10, is hard to model as the model needs to take into account the number of possible paths from each starting point. (\textbf{Middle}). Empirical density (\textbf{Right}). Learned density. The model is capable of modeling the density in two directions when conditioned on one position even if one order is more complicated than the other. }
\end{figure}

When trained with a left-to-right order, the model quickly learns the lazy random walk distribution but is never asked to learn more than that. However, when trained with a random order, the model has to learn to compute the more complex multinomial distribution, which is a much harder task as the model has to learn the global statistics of the sequence.

\subsection{Learning a harder problem: Chain rule of probability} \label{chain-rule}

Shuffling the order of the tokens in a sequence has the effect of creating a harder problem for the model. But if the model has access to enough data and to learn the distribution perfectly then shuffling should not change the results.
If the model learned the final distribution perfectly, the decoding order should not matter in theory as stated by the chain rule, indeed

\begin{multline}
  p(X_1 = x_1,X_2 = x_2,\dots, X_T = x_T) =  \\
  p(X_1 = x_1)  p(X_2 = x_2 \vert X_1 = x_1) \dots \\
  p(X_T = x_t \vert X_1 = x_1, \dots X_{T - 1} = x_{t - 1}) \\
  = p(X_{\sigma_1} = x_{\sigma_1})  p(X_{\sigma_2} = x_{\sigma_2} \vert X_{\sigma_1} = x_{\sigma_1}) \dots \\
  p(X_{\sigma_T} = x_{\sigma_t} \vert X_{\sigma_1} = x_{\sigma_1}, \dots X_{\sigma_{T - 1}} = x_{\sigma_{t - 1}})
\end{multline}

For any permutation $\sigma = \begin{pmatrix}
    1        & 2\,      & \cdots & T        \\
    \sigma_1 & \sigma_2 & \cdots & \sigma_T
  \end{pmatrix}$.

This is not what we see in practice, as models are not perfect and datasets are finite, so the impact of the order has its importance.

Another demonstration of this effect is the chain of thought step-by-step prompting strategy~\cite{wei2022cot,kojima2022stepbystep} of large language models. Where adding a few tokens between the question and the answer increases the working memory of the transformer, which seems to help the models make fewer mistakes.

\section{Estimation of number of steps for burst sampling} \label{burst-sampling}

\subsection{Deterministic case and lucky samplings}

In this section, we describe the fully deterministic case and the lucky samplings. In the deterministic case, and assuming a perfect model, the model will accept the sequence in one step. Indeed, if the model already knows the sequence, the probability distribution won't change when it is seen under a different order, therefore the rejection sampling ratio will always be one and the model will accept the sequence in one step.

In is the same for lucky samplings, again assuming a perfect model and a valid sample, the probability of a token given a specific order can only be higher than the probability of the token conditioned only on the prompt, as the tokens seen can only remove valid possibilities.
Hence $q(x) > p(x)$ and the model will accept the sequence in one step.

\subsection{Product Dataset: Number of passes needed to generate a valid sequence} \label{product-dataset}

In this section, we give the number of steps required to generate a valid sequence in the case of the product dataset using our rejection sampling scheme under a perfect model. The product dataset is made of a sequence of length 100 with two classes (0,1) given by a Bernoulli law with p = 10\%. At the beginning of the generation, a perfect model will predict $p(x_t=1|\{\}) = 0.1, \forall t \in \{1,\dots, 100 \}$. As all tokens are independent, the conditioning of the model will not change the probability of the next token. Therefore, the model will accept all tokens in the first step of the rejection sampling scheme. Our scheme should generate a valid sample for this distribution in at most 1 step.

\subsection{Step Dataset: Number of passes needed to generate a valid sequence} \label{step-dataset}

In this section, we give the number of steps required to generate a valid sequence in the case of the step dataset using our rejection sampling scheme under a perfect model. The step dataset is made of a sequence of length 100 of classes $\{0,1\}$ The sequence is 0 everywhere except on a step of ones of length 10 placed randomly in the sequence. At the beginning of the generation, a perfect model cannot do better than predicting $p(x_t=1|\{\}) = 0.1, \forall t \in \{1,\dots, 100 \}$.

Sampling with this law will lead to a sequence with 10 ones on average. The number of accepted tokens in the first round of rejection sampling will then depend on the random order sampled. The first token given by the order will always be accepted as $q(x_t|\{\}) = p(x_t|\{\})$. Now the next tokens are either compatible with the tokens already accepted or not. If the next token is compatible with the already accepted tokens, then it will be accepted with probability 1 as $p(x_t|\{\}) = q(x_t|x_{\text{accepted}})$. If it is not compatible, then it will be rejected as $q(x_t|\{x_{\text{accepted}}\}) = 0$ and the rejection sampling stops rejecting all the remaining tokens.

Now the reason for non-compatibility in the step dataset is either that

a. the model has already seen a 1, and sees a second 1 that is too far away,

b. that the model sees a one and then a zero nearby which would mean the end of the sequence which is not long enough, or

c. that it has seen only 0s, and no places are remaining for 1s. Therefore it has to reject the zero that would lead to an empty sequence. 1. In case, a-b. the model accepts a one, in the first step, and in case c. the start of the step is confined between the two last zeros where there are still enough places. 2. Once a token from the step is accepted, or there is only one place for a step, the signal is deterministic and a perfect model will 3. sample a complete step,  Our scheme should generate a valid sample for this distribution in at most 3 steps.

\subsection{Permutation}

When wanting to generate a valid permutation per burst assuming a perfect model, when given a partially completed permutation, a perfect model cannot do better than predicting a uniform law on the remaining tokens.

There is a well-known formula for the number of different classes that one gets when sampling with a uniform law, it is given by:
\begin{align}
  \E & [\text{\# of incoherent tokens}]         \\
     & = T- \E[\text{\# classes being sampled}] \\
     & = (T - 1)^ T / T ^{T - 1}
\end{align}

More generally, the probability of sampling exactly k classes is given by
\begin{align} \label{eq:permutation-probability}
  p & (\text{sampling exactly } k \text{ classes}) = \binom{T}{k} k! \begin{Bmatrix}T \\ k\end{Bmatrix} / T^T \\
\end{align}

Where $\begin{Bmatrix}T \\ k\end{Bmatrix}$ is the Stirling number of the second kind.

Having a closed formula for the expected number of steps is hard, but we can estimate it numerically using \Cref{eq:permutation-probability}. For a sequence of length 100, we find that the expected number of steps is 5.2 with a standard deviation of 0.6. Which gives us a lower bound for the number of steps required to generate a valid permutation.

Note that in this specific case, for our rejection sampling scheme to be able to generate a valid permutation in this number of steps, the model needs to validate the partial sequence under an order which is such that each unique classes are seen first because once a class is repeated the model will have to reject the rest of the sequence.


\section{Caching Scheme for Burst rejection sampling} \label{kvcaching}

Autoregressive transformers usually rely on a KV-caching mechanism to go from a $\gO (N^3)$ to a $\gO (N^2)$ cost of generation.
this KV cache can be adapted to generate a sample by burst. If we already have accepted a set of tokens $T_1$ and we have a KV cache $K_1, V_1$ that stores the keys and values for this set of tokens, and we want to generate a prediction for all the remaining set of $T_2$ tokens in parallel.

The output of the transformer corresponding to the remaining set of tokens is

\begin{equation}
  V_2' = \softmax \left(
  \begin{array}{c|c}
    Q_{T_2} K_1^T & D
  \end{array}
  \right) \left(
  \begin{array}{c}
    V_1 \\ \hline
    V_2
  \end{array}
  \right)
\end{equation}

With $D = \mathrm{diag}(q_i \cdot v_i) $
This formula parallelizes easily. We give the corresponding code in pytorch.

\begin{lstlisting}[language=Python,caption=KV Caching scheme of burst sampling,captionpos=b]
def burst(self, kv_cache, q, k, v):
    k1, v1 = kv_cache.get()

    att1 = q @ k1.transpose(-2, -1)
    # no masking on KV cache, tokens can see them all

    att2 = (q * k).sum(-1, keepdim=True)
    att = torch.cat((att1, att2), dim=-1) * (1.0 / math.sqrt(k.size(-1)))
    att = F.softmax(att, dim=-1)
    att = self.attn_drop(att)

    att1 = att[..., :-1]
    att2 = att[..., [-1]]

    # Works even if the cache is empty
    y = att1 @ v1 + att2 * v
    y = rearrange(y, "b h t e -> b t (h e)")

    y = self.resid_drop(self.proj(y))

    return y
\end{lstlisting}

\section{Additional experiments for the vertical rate forecasting} \label{exp-additional-details}

We present here additional results concerning the vertical rate modeling experiment~\Cref{tab:complete-results}. We report the Mean Square Error (MSE) compared to the ground truth when prompted by partially completed trajectories. The idea is the see the effect of partial left-to-right conditioning on the quality of the generated sequences. 0\%, 10\%, and 50\% denote the percentage of the actual flight that is given as a prompt to the model. In this setup, we see that $\sigma$-GPT outperforms models trained causally for a generation. As the dataset size is small, we noticed that models trained in a left-to-right manner were suffering from a repetition problem that arises when the modeling capabilities of the models are insufficient. As the prompting is given to the model in left-to-right order, we see that causal models can outperform random models when prompted with half of the actual sequence.

We see as well that diffusion models usually outperform autoregressive models in this task. However, they need to be retrained to be able to generate sequences conditioned on a partial sequence. We see that the autoregressive models can be used in a more flexible way as they can be used to generate sequences conditioned on a partial sequence without retraining.

\begin{table*}
  \caption{
    This table presents the Mean Square Error (MSE) results for the climbing rate forecasting task. The MSE is calculated on the entire generated sequence, conditioned on different points during the climb: the start (0\%), early stage (10\% of the climb), midway (50\% of the climb) and in the middle of the first climb. For each validation sequence, we generated 20 sequences autoregressively using the model, following three different schemes: causal scheme, random scheme, and binary search tree order. We also report the performance of diffusion models, for comparison, but we only report their performance for the entire sequence as they need to be retrained to be able to generate sequences conditioned on a partial sequence.}
  \label{tab:complete-results}
  \centering
  \begin{tabular}{llrrrr}
    \toprule
    \bf{Size}                      & \bf{Method}    & \bf{0\%}              & \bf{10\%}            & \bf{50\%}            & \bf{Mid Climb}      \\ 
    \midrule
    \multirow[c]{3}{*}{\bf{Small}} & \bf{Fractal}   & 145.7 $\pm$ 2.6       & 1278.9 $\pm$ 161.9   & 1466.6 $\pm$ 172.2   & 623.6 $\pm$ 84.4    \\ 
                                   & \bf{Causal}    & 274.8 $\pm$ 70.7      & 179.9 $\pm$ 48.0     & \bf{52.9 $\pm$ 12.7} & 46.1 $\pm$ 10.0     \\ 
                                   & \bf{Random}    & 141.4 $\pm$ 4.1       & 123.8 $\pm$ 6.0      & 63.8 $\pm$ 1.2       & 40.1 $\pm$ 2.7      \\ 
                                   & \bf{Diffusion} & \bf{105.94 $\pm$ 1.3} & -                    & -                    & -                   \\ 
    \cmidrule(l){2-6}
    \multirow[c]{4}{*}{\bf{Base} } & \bf{Fractal}   & 158.1 $\pm$ 5.9       & 2523.5 $\pm$ 13.9    & 1623.5 $\pm$ 283.5   & 884.4 $\pm$ 15.7    \\ 
                                   & \bf{Causal}    & 424.7 $\pm$ 10.9      & 287.7 $\pm$ 0.5      & 70.0 $\pm$ 5.7       & 61.5 $\pm$ 6.8      \\ 
                                   & \bf{Random}    & 146.3 $\pm$ 6.4       & 119.3 $\pm$ 7.8      & 65.3 $\pm$ 5.1       & 41.6 $\pm$ 0.7      \\ 
                                   & \bf{Diffusion} & 107.96 $\pm$ 2.4      & -                    & -                    & -                   \\ 
    \cmidrule(l){2-6}
    \multirow[c]{3}{*}{\bf{Tiny} } & \bf{Fractal}   & 152.3 $\pm$ 8.7       & 2553.6 $\pm$ 168.3   & 1508.8 $\pm$ 220.3   & 972.8 $\pm$ 44.9    \\ 
                                   & \bf{Causal}    & 290.9 $\pm$ 13.0      & 224.0 $\pm$ 1.8      & 58.9 $\pm$ 1.6       & 53.9 $\pm$ 5.7      \\ 
                                   & \bf{Random}    & 129.5 $\pm$ 23.7      & \bf{108.1 $\pm$ 6.1} & 72.6 $\pm$ 4.7       & \bf{36.7 $\pm$ 0.8} \\ 
                                   & \bf{Diffusion} & 123.28 $\pm$ 6.9      & -                    & -                    & -                   \\ 

    \bottomrule
  \end{tabular}
\end{table*}

\end{document}